%% file: bare_jrnl_new_sample4.tex
\documentclass[lettersize,journal]{IEEEtran}
\usepackage{amsmath,amsfonts}
\usepackage{amsmath}
\usepackage{bm}
\usepackage{algorithmic}
\usepackage{algorithm}
\usepackage{array}
\usepackage[caption=false,font=normalsize,labelfont=sf,textfont=sf]{subfig}
\usepackage{textcomp}
\usepackage{stfloats}
\usepackage{url}
\usepackage{verbatim}
\usepackage{graphicx}
\usepackage{cite}
\usepackage[]{multirow}
\usepackage{xcolor}
\usepackage{xspace}
\usepackage{inputenc}
\begin{document}

\title{Deep Learning Driven Natural Languages Text to SQL Query Conversion: A Survey}

\author{Ayush Kumar, Parth Nagarkar,
Prabhav Nalhe, and Sanjeev Vijayakumar
\thanks{}}

\markboth{Journal of \LaTeX\ Class Files,~Vol.~14, No.~8, August~2021}%
{Shell \MakeLowercase{\textit{et al.}}: A Sample Article Using IEEEtran.cls for IEEE Journals}


\maketitle

\begin{abstract}
With the future striving toward data-centric decision-making, seamless access to databases is of utmost importance. There is extensive research on creating an efficient text-to-sql (TEXT2SQL) model to access data from the database. Using a Natural language is one of the best interfaces that can bridge the gap between the data and results by accessing the database efficiently, especially for non-technical users. It will open the doors and create tremendous interest among users who are well versed in technical skills or not very skilled in query languages. 
Even if numerous deep learning-based algorithms are proposed or studied, there still is very challenging to have a generic model to solve the data query issues using natural language in a real-work scenario. The reason is the use of different datasets in different studies, which comes with its limitations and assumptions. At the same time, we do lack a thorough understanding of these proposed models and their limitations with the specific dataset it is trained on. 
In this paper, we try to present a holistic overview of 24 recent neural network models studied in the last couple of years, including their architectures involving convolutional neural networks, recurrent neural networks, pointer networks, reinforcement learning, generative models, etc. We also give an overview of the 11 datasets that are widely used to train the models for TEXT2SQL technologies. We also discuss the future application possibilities of TEXT2SQL technologies for seamless data queries.

\end{abstract}


\input{sections/introduction}
\input{sections/taxonomy}

\input{sections/benchmark}

\input{sections/algorithms}

\input{sections/conclusion}

 

\bibliographystyle{IEEEtran}
\bibliography{bare_jrnl_new_sample4}

\newpage

\vfill

\end{document}

%% file: sections/introduction.tex
\section{Introduction}

\IEEEPARstart{I}{n} today's world, a considerable portion of data is saved in relational databases for applications ranging from finance and e-commerce to medicine. As a result, it is not surprising that utilizing natural language to query a database has a wide range of uses. It also opens up the possibility of self-serving dashboards and dynamic analytics, where individuals unfamiliar with the SQL language may utilize it to obtain the most relevant information for their organization. Many activities are associated with converting natural language to SQL~\cite{xu2018sqlnet,yu2018typesql,wang2020ratsql,hazoom2021text2sqlinwild}, including code creation and schema construction. However, creating SQL is more complex than the typical semantic parsing problem. A brief natural language inquiry may necessitate combining numerous tables or having multiple filtering requirements. This needs more context-based techniques. In recent years, with the widespread development of deep learning techniques, particularly convolutions and recurrent neural networks, the outcomes have drastically improved for this purpose.

\begin{figure}
\centering
\includegraphics[width=0.49\textwidth]{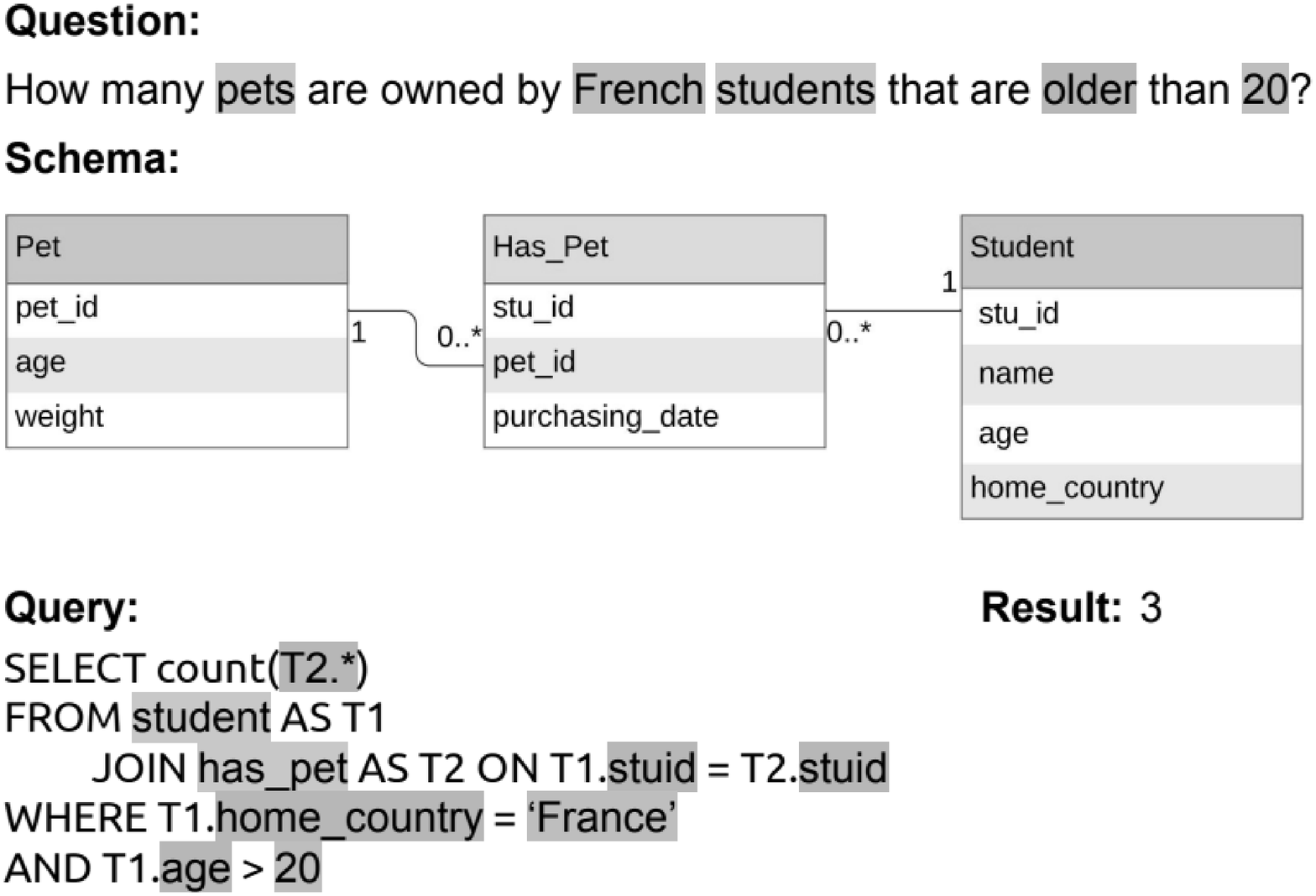}
\caption{ A generic NLP TEXT2SQL schema that generated a SQL query
given a natural language question and a database schema~\cite{brunner2021valuenet}. Other than understanding the requirement for a valid SQL query (non-highlighted words), the NLP algorithm also has to select the correct tables, columns, and values from where it will fetch the result}
\label{fig:valuenet}
\end{figure}

In general TEXT2SQL algorithms involve converting a natural language statement or texts into SQL query~\cite{semantic1,semantic2,semantic3,semantic4,semantic5} to access the desired dataset from the respective database. The input in the form of natural language text from the users is fed through various TEXT2SQL algorithms using baseline algorithms such as convolutional neural networks, recurrent neural networks, pointer networks, reinforcement learning, and generative model. Using these algorithms and input text queries, desired SQL queries are generated by concatenating various conditions used in the input. These SQL queries are then used to access the required dataset from the respective databases or a combination of various databases for many applications. It might eventually be integrated to make a broader goal of translating natural language into a fully functional application, combining with different forms or visualizations~\cite{Kumar_ETRA:2019,kumar2020visualthesis,kumar2018visual} or visual analytic tools~\cite{kumar2021eyefix,kumar2020demoeyesac}. One such sample for data query from natural language is shown in Figure~\ref{fig:valuenet} by Brunner et al.~\cite{brunner2021valuenet}.

The availability of massive annotated datasets containing questions and database queries has speeded up and caused substantial progress in the field by allowing the construction and deployment of supervised learning models for the job. This feat has been achieved by improving the test sets’ accuracy given with the datasets and focusing on developing the problem formulation toward higher complications more closely approaching applications that can be used in the real world. New datasets such as WikiSQL~\cite{wikisql2017} and Spider~\cite{spider18} pose the real-life challenge of generalization to unseen database schemas, where Stack Exchange data~\cite{hazoom2021text2sqlinwild} is taken from real-world example itself, which makes it equally challenging. There is a multi-database schema to which each query is mapped, and there is no overlap among the databases between the training and test sets. Multiple reasons make the generalization of schema challenging. 

There are multiple annotated datasets used for the Text2SQL~\cite{hazoom2021text2sqlinwild} task. 
The most commonly used datasets are Spider~\cite{spider18} and WikiSQL~\cite{wikisql2017}. Spider is a large-scale, complex, cross-domain semantic parsing and text-to-SQL dataset annotated by 11 college students. It comprises 10,181 questions and 5,693 distinct complex SQL queries on 200 databases with numerous tables spanning 138 domains. The WikiSQL corpus contains 87,726 hand-annotated SQL queries and natural language question pairings. These SQL queries are further classified as training (61297 examples), development (9145 examples), and test sets (17,284 examples). It may be used for relational database-related natural language inference problems. Multiple approaches have been used for the Text2SQL task, such as semantic parsing, IRNET, RuleSQL, etc., which use sophisticated methods to solve the Text2SQL task and reduce the error rate.

This overview paper presents comprehensive research on the most often used $11$ datasets, as shown in Table~\ref{tab:benchmark-table}, and 24 most recent (2018-2022) algorithms used on these datasets to address the challenge of synthesizing SQL queries from natural language texts. We looked into all the major NLP conferences such as ACL, EMNLP, IJCNLP, SIGKDD, ICLR, Computational Linguistics, and many other conferences and journals. The primary goal of this work is to offer a complete description and analysis of the most recent approaches for dealing with the issue of producing SQL using natural language, as well as the many datasets and evaluation methodologies that are constantly being improved.

%% file: sections/taxonomy.tex
\section{Terminology}


\textbf{LSTM:} 
Long Short-Term Memory (LSTM)~\cite{lstm1997} networks are a type of recurrent neural network (RNN)~\cite{seq2seq} capable of gauging the dependence of order in problems that require the prediction of sequential data. They are required in convoluted problem domains like machine translation, recognition of speech, etc. Long Short Term Memory Networks are well-suited to analyzing, processing, and predicting based on data in the time-series format. This is because many times, there are lags of unrevealed time scales between two or more events that might be significant. 

\textbf{Encoder:} Encoder~\cite{cho2014learningencoderdeco2, sutskever2014sequenceencoderdeco1} is a pile of multiple recurrent units such as LSTM, where each gets input in the form of a single element of the input sequence, gathering data for that specific element and generating it forward. It follows a procedure to transform the relevant text into number/vector representation to conserve the conditions and connection between words and sentences, such that a machine can comprehend the pattern associated with any text and make out the context of the sentences. 

\textbf{Decoder:} Decoder~\cite{cho2014learningencoderdeco2, sutskever2014sequenceencoderdeco1} is a pile of multiple recurrent units in which an output y for every time step is predicted. The current recurrent unit accepts a hidden state from the earlier recurrent unit. This is generally used in question-answering problems where the sequence of outputs is a collection of answers’ words. 

\textbf{BERT:} BERT~\cite{devlin2019bert} is an open source machine learning framework for natural language processing (NLP). It is designed to assist machines with comprehending vague language in the text by using nearby text to organize context. A Transformer is used along with the attention mechanism~\cite{vaswani2017attention} to learn the relations with respect to their context between multiple words in a sentence.

\textbf{Semantic Parsing:}
Semantic Parsing~\cite{krishnamurthy2017neuralsemanticpars} is used for transforming a natural language like English into a form that machines can comprehend in a logical form~\cite{semantic1,semantic2,semantic3, semantic4, semantic5}. The transformed languages can include SQL or any other conceptual representations.

\textbf{Baseline Model:}
A baseline model is a model that is very simple to implement or configure and, in most cases, provides satisfactory results. Trying to run any experiments with baseline models is fast and requires minimal cost. Usually, researchers use baseline models and leverage them to try to describe how their own trained model is better. The score of the baseline model is generally kept as a threshold.




\textbf{Incremental Decoding:}
Incremental decoding~\cite{huang2010efficientincrementaldecod} is a method in which translations are done dynamically as a set of input is parsed to the model. This approach refrains from waiting for the entirety of input to be parsed; instead produces output instantaneously as input is received. 

\textbf{Exacting matching accuracy:} It is the percentage of questions that outputs the exact SQL query as expected.

\textbf{Component matching F1:}It is the cumulative F1 scores.


%% file: sections/benchmark.tex
\section{Benchmark Dataset}

The datasets designed for semantic parsing of natural language phrases to SQL
queries consist of a combination of Annotated complicated questions and SQL queries. The
sentences are inquiries for a certain area. In contrast, the answers come from existing databases
resulting in a link between the question to its corresponding SQL query such that on the
execution of the SQL query, the result is obtained from existing databases. Several semantic
parsing datasets for SQL query been generated recently, each with different yet
significant characteristics. Table 1 from Kalajdjieski et.al~\cite{kalajdjieski2020recentbenchmarkdataset} contains extensive statistics of the most popular datasets
utilized by researchers. ATIS~\cite{atis-3}, GeoQuery~\cite{tang2001usinggeoquery}, Restaurants~\cite{tang2000automatedrestaurant1}, Academic~\cite{li2014constructingacademicdata}, IMDB~\cite{imdb2015}, Scholar~\cite{iyer2017learningscholardata},
Yelp~\cite{yaghmazadeh2017sqlizerYelpdata}, Stack Exchange Data~\cite{kalajdjieski2020recentbenchmarkdataset}, and Advising~\cite{finegan2018improvingadvisingdata} are examples of early datasets that focused on a single topic and
database. WikiSQL2~\cite{wikisql2017} and Spider~\cite{spider18} are newer datasets that are context-agnostic and greater
in size across domains. Additionally, new datasets include more queries and in-depth research, thus catering to effective model evaluation. To assess the model generalization
capabilities, previously unknown advanced queries may be used in the test sets. The authors of
Advising demonstrate that standard data splits increase the generalizability of the systems.
Even though the WikiSQL dataset comprises many questions and SQL queries, the SQL
queries are short and limited to certain tables. On the other hand, WikiSQL consists of more questions and SQL queries than the Spider dataset, which is composed of questions
that include various SQL expressions such as table join and nested query, making them of
added complexity in comparison. Spider dataset extensions SParC~\cite{yu2019sparc} and CoSQL~\cite{yu2019cosql} are developed
for contextual cross-domain semantical parsing and conversational dialog text-to-SQL systems.
As a result, these fresh aspects provide new and significant issues for future research in this
sector.

\textbf{ATIS (Air Travel Information System) / GeoQuery:}
ATIS is also known as Airline Travel Information Systems, a dataset~\cite{atis-3} consisting of audio recordings and hand transcripts of individuals utilizing automated travel inquiry systems seeking information about flights. The data is categorized into main purpose categories where the train, development, and test sets are composed of 4478, 500, and 893 intent-labeled reference utterances, respectively. ATIS, most commonly used for semantic parsing, is a methodology for facilitating the conversion of natural language queries into a formal meaning representation. On the other hand, GeoQuery consists of seven tables from the US geography database and 880 natural language to SQL pairings. Unlike WikiSQL, all queries in ATIS and GeoQuery are on a single domain. Resultantly utilizing them to train a deep learning model makes the model work only in one domain. Both these benchmarks contain varied queries with join and nested queries inclusive, whereas GeoQuery possesses a grouping and ordering query that the latter, i.e., ATIS, does not. 

\textbf{IMDb:} 
The IMDb dataset~\cite{imdb2015} is a massive collection of 50K IMDb reviews. Each film is limited to 30 reviews. There are an equal amount of favorable and negative reviews in the dataset. The dataset developers took into account extremely polarized reviews, with a negative review receiving a score of 4 out of 10 and a favorable review receiving a score of 7 out of 10.
When constructing the dataset, neural reviews are not taken into account.
The dataset is distributed evenly between training and testing.

\textbf{Advising:}
The advising dataset~\cite{finegan2018improvingadvisingdata} was intended to suggest improvements to text2SQL systems. The dataset's designers compare human-authored versus computer-produced inquiries, noting query features relevant to real-world applications. The dataset comprises university students' questions concerning courses that lead to complicated queries. The database contains fictitious student records. The dataset consists of student profile information such as suggested courses, grades, and introductory courses taken by the student. Students with knowledge of the database developed inquiries and were instructed to frame queries they may ask in an academic advising appointment. Many searches in the dataset were similar to those in the ATIS, GeoQuery, and Scholar databases.

\textbf{MAS :} 
Microsoft Academic Search includes a database of academic and social networks and a collection of queries. MAS~\cite{roy2013microsoftMAS}, like ATIS and GeoQuery, operates on a single domain. It has 17 tables in its database and 196 natural languages to SQL pairs. MAS contains a variety of SQL queries that include join, grouping, and nested queries but no ordering queries. The following limitations apply to each natural language query in MAS. To begin, a natural language question starts with "return me." In real-world scenarios, a user may inquire using an interrogative statement or a collection of keywords; however, MAS does not include such cases. Second, each natural language inquiry follows proper grammatical conventions.

\textbf{Spider :} 
Spider~\cite{spider18} is a large-scale, complicated, cross-domain semantic parsing and text-to-SQL dataset that 11 Yale students annotated. The Spider challenge's purpose is to provide natural language interfaces to cross-domain databases. It comprises 10,181 questions and 5,693 distinct sophisticated SQL queries on 200 databases with numerous tables across 138 domains. In Spider 1.0, train and test sets contain a variety of complicated SQL queries and databases. To perform effectively, systems must generalize not just to new SQL queries but also to new database structures.

\textbf{WikiSQL :}
WikiSQL~\cite{wikisql2017} is the most popular and most extensive benchmark, with 24,241 tables and 80,654 natural languages to SQL pairings in a single table. Tables are taken from Wikipedia's HTML tables. Then, for a given table, each SQL query is automatically created under the constraint that the query yields a non-empty result set. Each natural language inquiry is made using a basic template and then paraphrased by Amazon Mechanical Turk. All SQL queries in WikiSQL follow the same syntactic pattern: SELECT FROM T [WHERE (and)*], where T is a single table. This allows for a single projected column and conjunction option. It is worth noting that this syntax conveys no grouping, sorting, join, or nested queries.

The significant distinction between WikiSQL and Spider is that SQL queries in Spider are more complicated than those in WikiSQL. Table 1 is a complex SQL example from Spider in which the query appears theoretically easy but contains multiple elements of database structure and SQL clause. Aside from that, the Spider database has many tables, but the WikiSQL database has only one. The presence of many tables adds column and table name disambiguation issues to Spider, which do not exist in WikiSQL.

While WikiSQL and Spider are cross-domain settings, most SQL queries do not need domain expertise during the creation process. Domain knowledge is a consensus on just one issue and will not be articulated clearly in the inquiry. For example, when domain knowledge is necessary, asking for a 'good restaurant' might correspond to a WHERE condition 'star' greater than $3.5$ since this domain assesses places from 0 to 5 stars? Some domain examples replace words associated with schema item names while keeping the same phrase structure. Furthermore, rather than utilizing synonyms, most sentences use words directly connected to schema item names, allowing the model to discover the schema items using word matching.

\textbf{Stack Exchange :}
Stack Exchange Data Explorer (SEDE)~\cite{hazoom2021text2sqlinwild}
SEDE is an online question and answers community with over 3 million questions; it recently released a benchmark dataset of SQL queries consisting of  29 tables and 211 columns. This dataset is collected from real usage on the Stack Exchange website of common utterance topics such as published posts, comments, votes, tags, awards, etc. These datasets are also challenging to handle while semantically parsing as they have various real-world questions. We show that these pairs contain a variety of real-world challenges, which were rarely reflected so far in any other semantic parsing dataset. 1,714 questions out of 12,023 questions (clean) asked on the platform are verified by humans, making it an excellent choice for validation and test set while training the model.

\begin{table*}[h!]
\begin{center}
\begin{tabular}{ |c | c | c |c |c |c |}
\hline
 Dataset & Year & Tables & Questions & Unique Queries & Domain \\ 
 \hline
 \hline
 ATIS~\cite{atis-3} & 1994 & 25 & 5280 & 947 & Air Travel Information System \\  
 \hline
 GeoQuery~\cite{tang2001usinggeoquery} & 2001 & 8 & 877 & 246 & US geography database \\
 \hline
 Restaurants~\cite{tang2000automatedrestaurant1} & 2000 & 3 & 378 & 23 & Restaurants, Food Type, Locations \\
 \hline
 Academic~\cite{li2014constructingacademicdata} & 2014 & 15 & 196 & 185 & Microsoft Academic Search \\
 \hline
 IMDB~\cite{imdb2015} & 2015 & 16 & 131 & 89 & Internet Movie Database \\
 \hline
 Scholar~\cite{iyer2017learningscholardata} & 2017 & 7 & 817 & 193 & Academic Publications \\
 \hline
 Yelp~\cite{yaghmazadeh2017sqlizerYelpdata} & 2017 & 7 & 128 & 110 & Yelp Movie Website \\
 \hline
 WikiSQL~\cite{wikisql2017} & 2017 & 24241 & 80654 & 77840 & Wikipedia \\
 \hline
 Advising~\cite{finegan2018improvingadvisingdata} & 2018 & 10 & 4570 & 211 & Student Course Information \\
 \hline
 Spider~\cite{spider18} & 2018 & 645 & 10181 & 5693 & 138 Different Domains \\
 \hline
  SEDE~\cite{hazoom2021text2sqlinwild} & 2021 & 29 &  12,023 & 11767 & Stack Exchange \\
 \hline
\end{tabular}
\end{center}
\caption{\label{tab:benchmark-table}Tabulation of Benchmark Datasets~\cite{kalajdjieski2020recentbenchmarkdataset, hazoom2021text2sqlinwild}}
\end{table*}

%% file: sections/algorithms.tex
\section{Algorithmic explanation}
In this section, we focus on explaining 24 recent (2018 -2022) Text2SQL algorithms that introduce an improvement or adapt existing text2SQL techniques to address all common challenges researchers face.

\subsection{\textbf{SQLNet}~\cite{xu2018sqlnet}}
Synthesizing SQL queries from natural language has been a long-standing open subject lately garnered significant interest. This paper suggests a unique solution, SQLNet, to fundamentally tackle the problem of synthesizing SQL queries by bypassing the sequence-to-sequence structure where the order is entirely irrelevant. They use a sketch-based technique in particular, where the sketch incorporates a dependency network, such that a single prediction may be made by considering just the prior forecasts on which it is dependent. Additionally, this paper presents a sequence-to-set model and column attention technique for generating the query from the sketch. By integrating all of these unique strategies, the authors have demonstrated that SQLNet outperforms the previous state of state-of-the-art SQL challenge by 9\% to 13\%.

To achieve this, the authors have used a sketch strongly aligned with the SQL language. The sketch is intended to be broad enough that it may be used to represent any SQL queries that interest the developer. The sketch illustrates the interdependence of the predictions to be made. The approaches are combined to create an SQLNet neural network, which may be used to synthesize a SQL query from a natural language inquiry and a database table structure. The SQLNet and training specifics are also mentioned to outperform the prior state-of-the-art technique without using reinforcement learning.

\begin{figure}[h]
    \includegraphics[width=0.49\textwidth]{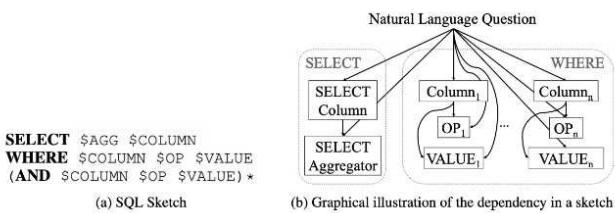}
    \caption{Sketch Syntax and dependency in a sketch~\cite{xu2018sqlnet}}
\label{arimaresults}
\end{figure}

\subsubsection{Sketch Based query synthesis}
A box represents each slot whose value is to be predicted, and a directed edge represents the relationship between each slot and its dependencies. For example, the box of OP1 has two incoming edges, one from Column1 and the other from the natural language query. These edges suggest that the prediction of the value for OP1 is influenced by both the values of Column1 and the natural language query provided. This dependency network serves as the foundation for Their model, which They may consider as a graphical model and the query synthesis issue as an inference problem based on the graph. When They look at it from this viewpoint, they can see that the prediction of one constraint is entirely independent of the prediction of another constraint. Therefore, Their technique can altogether "order matters" problem in a sequence-to-sequence model.

\subsubsection{Sequence-To-Set Prediction Using Column Attention}
Sequence-to-set: From an intuitive standpoint, the column names appearing in the WHERE clause represent a subset of the total number of column names present in the database. Consequently, rather than creating a series of column names, they can anticipate which column names will appear in this subset of interest, saving time and effort. This concept is referred to as "sequence-to-set prediction."
Notably, in this case, they compute the probability Pwherecol(col|Q), where col is the column’s name, and Q is the query in plain language. One approach to achieving this goal is to compute Pwherecol(col|Q) as where uc and uq are two column vectors of trainable variables, Ecol and EQ are the embeddings of the column name and the natural language question, and uc and uq are the embeddings of the column name and the natural language question, respectively.
When the sequences of col and Q are used as inputs to an LSTM, it is possible to calculate the embeddings Ecol and EQ by computing the hidden states of a bi-directional LSTM running on top of each sequence.

For each token in the query, they compute the attention weights w for that token. Specifically, w is an L-dimension column vector, which can be represented mathematically as

Column attention: They design the column attention mechanism to compute EQ|col instead of EQ. In particular, they assume HQ is a matrix of d×L, where L is the length of the natural language question. The i-th column of HQ represents the hidden state’s output of the LSTM corresponding to the i-th token of the question.n

\subsubsection{SQLNet Model and Training Details}
Predicting the WHERE clause:
Column Slots: SQLNet must determine which columns should be included in the WHERE clause. An alternative strategy is establishing a threshold of 0 or 1 such that all columns with Pwherecol(col|Q) 0 or 1 are picked. They see that the WHERE clauses of most searches contain just a small number of columns, as They will see below. As a result, they establish an upper bound of N on the number of columns to pick from, and They formulate the issue of predicting the number of columns as an (N + 1)-way classification problem to begin with (from 0 to N).

Predicting The Select Clause:
The SELECT clause contains an aggregator as well as a column designation. The prediction of the column name in the SELECT clause is similar to the prediction in the WHERE clause. The most significant distinction is that with the Choose clause, they only need to select one column from among all of the other possible choices. As a result, they calculate

Training Details
To parse the statement, the Stanford CoreNLP tokenizer is employed. To accomplish this, they use the GloVe word embedding~\cite{pennington2014glove} to represent each token as a one-hot vector.

\subsection{\textbf{Slot-Filling Approach}~\cite{goo2018slotfilling}}

In concordance with the name, a slot-filling approach imbibes the meaning of the sentence by a divide and conquer methodology~\cite{goo2018slotfilling}. Individual words encountered in the human language query are categorized into sub-domain,s also known as slots. With reference from the model TypeSQL~\cite{yu2018typesql}, the essential and sufficient slots are declared to be \$AGG, \$SELECT\_COL under the SELECT clause, which signifies the aggregation and selected column, allowing to capture the data required from a particular column and \$COND\_COL, \$OP, \$COND\_VAL under the WHERE clause and AND clause (if needed) which assists in filtering the data being extracted according to the user. The three slots aforementioned under the WHERE and AND clause signify the column subjected to conditioning, an operator, and the condition value, respectively. To fit individual words in these slots, an input encoder comprising two bi-directional LSTMs~\cite{graves2005bidirectionallstm} is utilized. The encoder classifies words based on predefined scilicet INTEGER, FLOAT, DATE, YEAR, and NAMED ENTITIES such as PERSON, PLACE, COUNTRY, ORGANIZATION, AND SPORT. The LSTMs are programmed accordingly to iterate over each column. After extensive research, the model was fabricated to work efficaciously with a sum of 3 models. 

\begin{table*}[h!]
\begin{center}
\begin{tabular}{ | c | c | c | c | c | c | c | }
\hline
\multicolumn{1}{|c|}{} & \multicolumn{3}{|c|}{Dev} & \multicolumn{3}{|c|}{Test} \\
\hline
\hline
\multicolumn{7}{|c|}{Content Insensitive} \\
\hline
    & $Acc_{If}$ & $Acc_{qm}$ & $Acc_{ex}$ & $Acc_{If}$ & $Acc_{qm}$ & $Acc_{ex}$ \\
  \hline
  Dong and Lapata (2016) & 23.3\% & - & 37\% & 23.4\% & - & 35.9\% \\
  \hline
  Augmented Pointer Network (Zhong et al., 2017) & 44.1\% & - & 53.8\% & 42.8\% & - & 52.8\% \\
  \hline
  Seq2SQL (Zhong et al., 2017) & 49.5\% & - & 60.8\% & 48.3\% & - & 59.4\% \\
  \hline
  SQLNet (Xu et al., 2017) & - & 63.2\% & 69.8\% & - & 61.3\% & 68.0\% \\
  \hline
  TypeSQL w/o type-awareness (self-made) & - & 66.5\% & 72.8\% & - & 64.9\% & 71.7\% \\
  \hline
  TypeSQL (self-made) & - & 68.0\% & 74.5\% & - & 66.7\% & 73.5\% \\
  \hline
  \multicolumn{7}{|c|}{Content Sensitive} \\
  \hline
  Wang et al. (2017a) & 59.6\% & - & 65.2\% & 59.5\% & - & 65.1\% \\
  \hline
  TypeSQL+TC (self-made) & - & 79.2\% & 85.5\% & - & 75.4\% & 82.6\% \\
  \hline
\end{tabular}
\end{center}
\caption{\label{tab:table-typesql}Experimental Result of Slot Filling Approach - TypeSQL derived from paper~\cite{yu2018typesql}}
\end{table*}

There are three evaluation metrics utilized to display the performance of the model, namely accuracies of canonical representation matches on AGGREGATOR, SELECT COLUMN, and WHERE CLAUSES depicted by Acc\_{If}, Acc\_{qm} and Acc\_{ex} respectively. The model outperforms the baseline models by 5.5\% based on executing accuracy—similarly, the accuracy based on SELECT and WHERE clauses are improved by 1.3\% and 5.9\%, respectively. When given complete access to the database, the model achieves 82.6\% based on executing accuracy and 17.5\% on the content-aware system. The results have been summarized in table \ref{tab:table-typesql}.

\subsection{\textbf{Sequence-to-Tree model}~\cite{cineseSQL2019pilot}}
This model's prime goal is that this approach's output is returned in the form of a tree. Considering the intricacy of returning a visual tree and its limitations in parsing its structure, a pre-order traversal of the tree is returned. The model uses a neural semantic parsing method of Yu
et al.~\cite{spider18} as the baseline model. CHISP~\cite{cineseSQL2019pilot}, parses the input question by encoding in a LSTM sequence encoder. After parsing the input, the model understands the input by recognizing individual entities and classifies information based on three categories. As it is essential and adequate for a SQL query to have data for SELECT, WHERE, EXCEPT, the model follows a similar basket approach. Ultimately, the tree nodes comprise keyword nodes as described above and columns of tables such as Name, City, etc.

In addition to this, a stack is utilized for incremental decoding, as explained in section 2.2. As log data is generated from previous accumulations of results, the model uses this data to decide the term that is supposed to be developed in the next iteration.

Exacting match accuracy and component matching F1 are used for evaluation for SELECT, GROUP BY, WHERE, and ORDER BY.

\begin{table}[h!]
\begin{center}
\begin{tabular}{ | c | c | c | c | c | c | c |  }
\hline
 \multicolumn{2}{|c|}{} & Easy & Medium & Hard & Extra Hard & All \\
 \hline
 \multicolumn{2}{|c|}{ENG} & 31.8\% & 11.3\% & 9.5\% & 2.7\% & 14.1\% \\
 \hline
 \hline
 \multirow{6}{*}{HT} & C-ML & 27.3\% & 9.9\% & 7.5\% & 2.3\% & 12.1\% \\
 & C-S & 23.1\% & 7.7\% & 6.2\% & 1.7\% & 9.9\% \\
 & WY-ML & 21.4\% & 8.1\% & 8.0\% & 1.7\% & 10.0\% \\
 & WY-S & 20.2\% & 6.4\% & 6.7\% & 2.0\% & 8.9\% \\
 & WJ-ML & 19.8\% & 8.6\% & 5.0\% & 1.3\% & 9.2\% \\
 & WJ-S & 20.1\% & 5.0\% & 5.7\% & 1.7\% & 8.2\% \\
 \hline
 \multirow{2}{*}{MT} & C-ML & 18.1\% & 4.6\% & 5.2\% & 0.3\% & 7.9\% \\
 & WY-ML & 17.9\% & 4.7\% & 4.5\% & 0.3\% & 7.6\% \\
 \hline
\end{tabular}
\end{center}
\caption{\label{tab:table-chisp}Experimental Result of Sequence to Tree Model~\cite{cineseSQL2019pilot}}
\end{table}

Summarizing of exact matching results has been depicted in table \ref{tab:table-chisp}. As it can be seen, The accuracy of each dataset under their respective categories has been populated.

\subsection{\textbf{Zero-Shot-Semantic Parsing}~\cite{bogin2019representingschema}}
This particular text to SQL parser has been extended from its predecessor known as the encoder-decoder semantic parser~\cite{bogin2019representingschema}. In an encoder-decoder~\cite{cho2014learningencoderdeco2} semantic parser, the training data is initially passed through a graph neural network~\cite{scarselli2008graphneuralnetwork} to initialize the DB constants in a soft selection manner. After this, an auto-regressive model is utilized to imbibe the top-K queries and rank the list based on global matching. Global matching in this setting signifies the process of considering words in the corpus in correspondence to the database rather than just focusing attention on the target word. The approach utilized in this model plays a significant part in improving the efficiency of the model when compared to the baseline papers.

To delineate the zero-shot-semantic parsing, the corresponding datasets and equations are employed. The training set resembles the following structure $(x^{k}, y^{k}, S^{k})$ where $x^{k}, y^{k}, S^{k}$ signifies a input question, translation to its corresponding SQL query, and the schema of the corresponding database.  $S^{k}$ in this scenario includes three important parameters: the set of DB tables, a set of columns for each table, and a set of foreign key-primary key column pairs wherein a pair represents a relation from a foreign-key table to a primary-key table. The input question, $x^i$, is encoded using a Bi-LSTM~\cite{graves2005bidirectionallstm} and the output query, $y^i$, is decoded at each step by decoding grammar using SQL grammar. 

\begin{equation}
\label{zeroshot-1}
    s_v = \Sigma_i \alpha_i s_{link} (v, x_i)
\end{equation}

\begin{equation}
\label{zeroshot-2}
    output = softmax([s_v]_{v \in V})
\end{equation}

\begin{equation}
\label{zeroshot-3}
    h_v^{(0)} = p_v · r_v
\end{equation}

To compute a DB constant, first attention scores~\cite{sun2020understandingattentionscore} are calculated as $\alpha^{|x|}_{i=1}$. Sequel to it, the local similarity score $s_{link}$ is computed from an edit distance between the two inputs, the fraction of string overlap between them, and the learned embeddings~\cite{elekes2020towardnlpembedding} of the word and DB constant. Proceeding further, equation \ref{zeroshot-1} is employed to compute $s_v$, and equation \ref{zeroshot-2} is used to circumscribe it with a softmax layer to obtain the desired output. Upon continuous research, it was concluded that a representation $h_v$ associated with every DB constant was used in the paper~\cite{bogin2019representingschema}. This representation is used in the prescribed Graph Convolutional Network (GCN)~\cite{gao2018largegraphCNN} in a node-to-node manner. Considering the excessive number of nodes present in one GCN, a probability factor known as relevance probability, $p_v$, was computed for every node and served as a gate for input to the network. The mathematical representation of the aforementioned has been depicted in equation \ref{zeroshot-3}. In addition, $p_v$ is derived based on local information rather than from a global perspective.

\begin{equation}
\label{zeroshot-3}
    g_v^{(0)} = FF([r_v; h_v; p_v])
\end{equation}

\begin{equation}
\label{zeroshot-4}
    p_{global} = \sigma(FF(g_v^{(L)})
\end{equation}

\begin{equation}
\label{zeroshot-5}
    f_{U_{\hat{y}}} = f^{(L_{v_{global}})}
\end{equation}

As previously suggested, the approach adopted utilizes a global gating methodology to allot the context of the question in contrast to its contemporary baselines. A new node $v_{global}$ is added to the GCN to accommodate this. $p_{global}$ is initialized randomly which is then evolved based on equation \ref{zeroshot-3} and equation \ref{zeroshot-4}. The FF in the equations \ref{zeroshot-3} and \ref{zeroshot-4} denotes the feedforward neural network and concatenation of the described columns. On a further note, to re-rank the desired output list, a quotidian approach is to use an autoregressive model to carry the job. The approach is set to train a separate discriminative model to account for the global superposition of words. Here, the sub-graph created by the node of interest $U_{\hat{y}}$, and the global node $v_{global}$ is fed as input according to the equation \ref{zeroshot-5}.

The model attains an accuracy of 47.4\%. The basis of experimentation is summed up by two broad domains when queries use a single table and multiple tables. Assuming a global gating increases the accuracy to 63.2\% in contrast to its contemporaries.

\subsection{\textbf{Zero-Shot-Semantic Parsing with induced dynamic gating}~\cite{chen2020talelinkingdynamic}}
As an evolved version of the zero-shot-semantic parsing aforementioned, this approach inherits an additional trait known as dynamic gating to fill a new entity in the GCN (Graph Convolutional Network). As already specified, a typical zero-shot-semantic parser can be classified into three the NL decoder that maps questions to word embeddings, a schema encoder that builds a relation-aware entity representation for every entity in the schema. Finally, a grammar decoder generates an Abstract Syntax Tree (AST)~\cite{zhang2019novelAST} using an autoregressive LSTM model. There arise two particular cases when the AST tree is about to generate:

\begin{itemize}
  \item to generate a new rule that is responsible for inducing the leftmost non-terminal node after each iteration
  \item to generate a DB schema according to a rule spawned by the above point in the previous iteration
\end{itemize}

There are two techniques been deciphered in the research study~\cite{chen2020talelinkingdynamic}, which are as follows:

\begin{itemize}
  \item Schema Linking: Compiles the output of NL encoder and chooses an apt entity based on string matching and embedding-matching.
  \item Structural Linking: Surfs through the generated entities and finds a perfect fit for the entity under consideration.
\end{itemize}

The whole structure of the aforementioned is already at play. The dynamic gating process assists in choosing between the two probabilities according to the situation. One denotes the likelihood of choosing Schema Linking, and the other represents the likelihood of selecting Structural Linking. The reason for a direct relation is because the required data lies with the desired entity and, in hindsight. Hence relying entirely on this parameter establishes control over the model.

\subsection{\textbf{Zero-shot Text-to-SQL Learning with Auxiliary Task}~\cite{chang2020zeroshot}}

This work begins by diagnosing the bottleneck in the text-to-SQL job by introducing an entirely novel testing environment where it is discovered that current models have weak generalization capacity on data that has only been viewed a few times. The authors are encouraged by the initial study’s results to develop a simple but effective auxiliary task that acts as both a supporting model and a regularization term to force generation jobs to boost the generality of the models.  On the overall dataset, this model outperforms a robust baseline coarse-to-fine model by more than 3\% in absolute accuracy compared to the baseline model. Interesting to note is that the model described by the author outperformed the baseline on a zero-shot subset test of WikiSQL, exhibiting a 5 percent absolute accuracy improvement over the baseline and proving its improved ability to generalize.

\begin{figure}[h]
    \includegraphics[width=3.5in, height=3in]{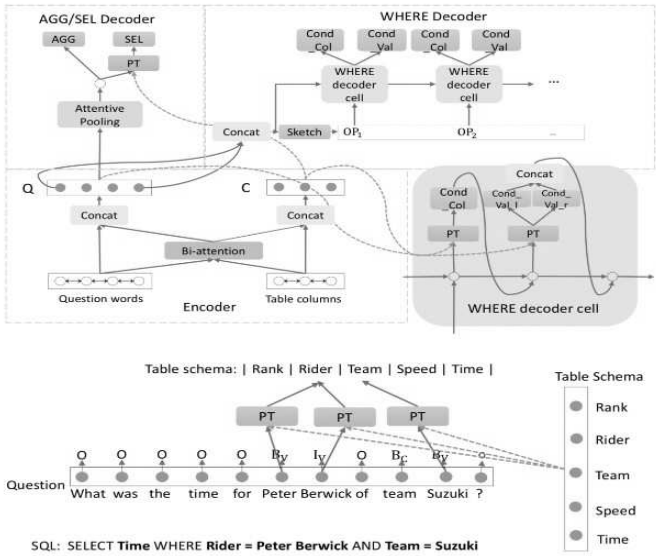}
    \caption{Zero-shot Text-to-SQL Learning with Auxiliary Task~\cite{chang2020zeroshot}}
\label{arimaresults}
\end{figure}

\subsubsection{Encoder}
The encoder retrieves the question's hidden representations and the schema, Hq, and Hc. For the final question and table schema representation, bi-attention is employed to strengthen the interaction between the question terms q and the column name c. The formula is: 

\subsubsection{AGG and SEL Decoder}
AGG and SEL are generated using an attentive pooling layer over H q, combined with an attentive pooling layer to provide a final hidden representation q SEL. They input qSEL into the CLS layer, which generates the aggregation operation AGG, and They assess the similarity score between qSEL and each column name C j, which allows the PT layer to forecast SEL in the following ways:

\subsubsection{WHERE Decoder}
To do this, They used the WHERE decoder from the current state-of-the-art model, which first creates a slot sketch of the WHERE clause before converting the SQL generation into a slot filling issue. WikiSQL has 35 categories of WHERE clauses, each of which is a subsequence of a WHERE clause that skips the COND COL and COND VAL. Examples include "WHERE = AND > ", a sketch of a WHERE clause that consists of two criteria. They begin by predicting the drawing q based on H q where qWhere = [h¯q1, h¯q |Q|].

\subsubsection{Auxiliary Mapping Model}
The three components of a SQL query condition are COND COL, COND OP, and COND VAL. Each state is composed of three parts. Its mapping model aims to figure out how to transfer the condition column to the condition value in a given situation. Hq and Hc are two representations of the question and table schema shared with the generation model. An intuitive mapping method is to immediately learn a mapping function from each word in question to the names of the columns in question. A two-step mapping model is proposed to handle this issue. First, a detector to filter out condition values is learned; then, a mapping function from condition values to column names is learned.

\subsection{\textbf{Air-Concierge}~\cite{chen2020airconcierge}}

AirConcierge, an end-to-end trainable text-to-SQL guided
framework to learn a neural agent that interacts
with KBs using the generated SQL queries.
Specifically, the neural agent first learns to ask
and confirm the customer’s intent during the
multi-turn interactions, then dynamically determining when to ground the user constraints into executable SQL queries to fetch relevant information from KBs.
AirConcierge system addresses the following challenges in developing an effective
task-oriented dialogue system, including
• When should the system access the KBs to
obtain task-relevant information during a conversation?
• How does the system formulate a query that
retrieves task-relevant data from the KBs

\subsubsection{Dialogue Encoder}
\label{airFirst}

\begin{equation}
\label{airconcierge-1}
   h_t^e = GRU(W_{emb}(x_t-1), h_{t-1}^e)
\end{equation}

The Dialogue Encoder is encoded using a Recurrent Neural Network (RNN)~\cite{cho2014learningGRU, chung2014empiricalgatedrnn}. Subsequent to embedding by the $W_{emb}$ matrix, equation \ref{airconcierge-1} is used to model the sequence of input conversation history X = \{x\_1, x\_2 , x\_3, ...x\_t\}

\subsubsection{Dialogue State Tracker}
\label{dialogue state tracker}

\begin{equation}
\label{airconcierge-2}
   P(s|h_T^e, x_{1:J}^{col}) = \sigma(W_2^s(W_1^sh_T^e + \Sigma U_2W_{emb}(x_{1:J}^{col})))
\end{equation}

Upon multiple iterations performed, there arises a situation where the model needs to determine if the data available with it is enough for the further steps or not. In detail, the system switches from the "greeting state" to the "problem-solving state". The dialogue state tracker is utilized to facilitate this task, wherein the schema from the database is taken as input data. Specifically, it takes the parameter as input and returns a binary value between 0 to 1 using bidirectional LSTM, with a fully-connected layers and a sigmoid function.

\subsubsection{SQL Generator}

\begin{equation}
\label{airconcierge-3}
   P_{col}(x_j^{col}|h_j^{col}, h_T^e) = \sigma(W_1^{col}h_j^{col} + W_2^{col}h_T^e)
\end{equation}

\begin{equation}
\label{airconcierge-4}
   P_{op}(x_j^{op}|h_j^{col}, h_T^e) = \sigma(W_1^{op}h_j^{col} + W_2^{op}h_T^e)
\end{equation}

\begin{equation}
\label{airconcierge-5}
   P_{val}(v_i^j|h_j^{col}, h_T^e) = Softmax(W_1^{val}(W_2^{val}h_T^e + W_3^{val}h_j^{col}))
\end{equation}

Once the model switches to the "problem-solving state", the SQL generator fires up to predict the values of three broad clauses, the COLUMN NAME, OPERATOR, and VALUE in the desired column in concordance with a SQL query. Equations \ref{airconcierge-3}, \ref{airconcierge-4}, and \ref{airconcierge-5} are used respectively to compute the values aforementioned.

\subsubsection{Knowledge base memory encoder}

\begin{equation}
\label{airconcierge-6}
  c_i^k = B(C^k(f_i))
\end{equation}

\begin{equation}
\label{airconcierge-7}
  p_i^k = Softmax((q^k)^T c_i^k)
\end{equation}

\begin{equation}
\label{airconcierge-8}
  o^k = \sum_{i=1}^{F} p_i^k c_i^{k+1} 
\end{equation}

\begin{equation}
\label{airconcierge-9}
  q^{k+1} = q^k + o^k
\end{equation}

\begin{equation}
\label{airconcierge-10}
  g_i^K = Softmax((q^K)^T c_i^K)
\end{equation}

In addition to encoding the retrieved data from the existing Knowledge base (KB), the knowledge base memory encoder filters out irrelevant data from the KBs. With the help of $C = \{ C^1, C^2,....C^{K+1}\}$, the trainable embedding matrices, the data retrieved from KBs are converted to memory vectors $\{ m_1, m_2....m_F\}$ where K is the number of hops. Simultaneously a vector q\^k, known as the attention weights, is computed for each memory vector m\_i. With the help of equations \ref{airconcierge-6} and \ref{airconcierge-7}, each memory vector's probability of higher relevance is calculated. 

Further, a parameter o\^k is calculated by equation \ref{airconcierge-8} which is used to update q\^K that can be visualized in equation \ref{airconcierge-9}. Finally, the vector/pointer G = (g\_1, g\_2,...g\_F) is used to pick out the most relevant data points amongst the heap and filter out non-essential data points. Individual entities of the pointer G can be calculated with the help of equation \ref{airconcierge-10}.

\subsubsection{Dialogue decoder}

\begin{equation}
\label{airconcierge-11}
 h_t^d = GRU(W_{emb}(\hat{y_t}-1), h_{t-1}^d)
\end{equation}

\begin{equation}
\label{airconcierge-12}
 P(\hat{y_t}) = Softmax((q^K)^T c_i^K)
\end{equation}

A dialogue decoder is utilized to decipher the agent’s output at every step. This is accomplished with the help of a GRU model for a seamless output. Equations \ref{airconcierge-11} and \ref{airconcierge-12} are responsible for the same.

\subsection{\textbf{Machine Reading Comprehension model}~\cite{yan2020sqlmachineread}}
The MRC (Machine Reading Comprehension) model~\cite{yan2020sqlmachineread} is based on the BERT-based MRC model that allows efficient conversion of text span in the model. It formulates the task as a question-answering problem. A unified MRC model is used to predict different slots with intermediate training on the dataset, achieving comparable performance on WikiSQL base state of the art.

\begin{equation}
\label{mrc-1}
 H^Q, H^C = BERT([Q, C])
\end{equation}

\begin{equation}
\label{mrc-2}
 p_{start}(i) = softmax(H_i^C v_{start})
\end{equation}

\begin{equation}
\label{mrc-3}
 p_{end}(i) = softmax(H_i^C v_{end})
\end{equation}

Initially, the question parsed to the model is denoted by Q, and the output/answer is concatenated to generate the training data. [CLS], q\_1, q\_2, .., q\_L, [SEP], c\_1, c\_2, ..., c\_M  would be an apt representation in a mathematical form, where [CLS] signifies the start and [SEP] is a unique character inserted to differentiate the question and the corresponding answer. As a result, the model outputs the matrix H. In this trial, two additional parameters, v \_{start} and v\_{end} wrapped under a softmax layer are used to compute the probability of the start and end token positions, respectively. The above model is suitable for predicting only one set of WHERE, OPERATOR, and VALUE. To overcome this problem, an alternate approach is enforced. 

\begin{equation}
\label{mrc-4}
 T = CRF(W^LH^C), |T| = |C|
\end{equation}

Sequence labeling using a BIO tag set is used for multiple sets of predictions. In this scenario, each token in the context, H\^C, is fed to a conditional random field to yield output labels T, as specified in equation \ref{mrc-4}. This approach is similar to an MRC-based entity-relation extraction and is under development. To enhance the efficiency of this study, an experimental approach was employed. The method known as STILTs involves utilizing a pre-trained model for intermediate tasks before the final deployed model.

\subsection{\textbf{Seq2SQL}~\cite{zhong2018seqsqlwithreinfolear}}
It is a deep neural network for translating questions about a natural language into their respective queries in the form of Structured Query Language. The developed model uses the formation of queries in Structured Query Language to reduce the output space concerning the generated questions remarkably. It is trained using a mixed objective, combining cross-entropy losses and RL rewards from in-the-loop query execution on a database.  Seq2SQL outperforms a previously state-of-the-art semantic parsing model on the WikiSQL benchmark dataset. It achieves 59.4\% execution accuracy compared to the previous 35.9\% and 53.3\% execution accuracy.

Concerning the model Sequence to SQL~\cite{dong2016languageatt_seq_seq}, the model is segregated into three individual aspects. The AGGREGATION operator, the SELECT column, and the WHERE clause

\subsubsection{AGGREGATION}

\begin{equation}
\label{seq2sql-1}
 \alpha_t^{inp} = W^{inp} h_t^{enc}
\end{equation}

\begin{equation}
\label{seq2sql-2}
 \alpha^{inp} = [\alpha_1^{inp}, \alpha_2^{inp},...]
\end{equation}

\begin{equation}
\label{seq2sql-3}
 \beta^{inp} = softmax(\alpha^{inp})
\end{equation}

\begin{equation}
\label{seq2sql-4}
 k^{agg} = \sum_{t} \beta_t^{inp} h_t^{enc} 
\end{equation}

\begin{equation}
\label{seq2sql-5}
 \alpha^{agg} = W^{agg}tanh(V^{agg} k^{agg} + b^{agg}) + c^{agg} 
\end{equation}

\begin{equation}
\label{seq2sql-6}
 \beta^{agg} = softmax(\alpha^{agg})
\end{equation}

Initially, the scalar attention scores are calculated according to equation \ref{seq2sql-1} which is further normalized as depicted in equation \ref{seq2sql-2}. The individual scores for the aggregation operators, COUNT, MIN, MAX and NULL can be calculated from equations \ref{seq2sql-3}, \ref{seq2sql-4}, and \ref{seq2sql-5}. Finally, a softmax layer is applied over the result to obtain the distribution over the possible aggregation operations.

\subsubsection{SELECT}

\begin{equation}
\label{seq2sql-7}
 h_{j, t}^c = LSTM(emb(x_{j, t}^c), h_{j, t-1}^c)
\end{equation}

\begin{equation}
\label{seq2sql-8}
 e_j^c = h_{j, T_j}^c
\end{equation}

\begin{equation}
\label{seq2sql-9}
 \alpha_j^{sel} = W^{sel} tanh(V^{sel}k^{sel} + V^c e_j^c)
\end{equation}

Each column names are encoded with a LSTM as shown in equation \ref{seq2sql-7}, where h\^c\_{j,t} denotes the t\^{th} encoder state of the j\^{th} column. Another representation k\^{sel} is fabricated using united weights, which is further used to wrap over a multi-layer perceptron over the column representations, to compute a score as shown in equation \ref{seq2sql-8} and \ref{seq2sql-9}. Finally, a softmax layer is applied over the output B\^{sel} to normalize the scores.

\subsubsection{WHERE}
To avoid the mismatch of queries, reinforcement learning is utilized to learn a policy to optimize and minimize the error produced. If an invalid query is parsed, the score is -2, whereas if the query is valid, but the query executes to an incorrect result, then the score is -1, and finally, for the perfect result, where the query parsed is right as well as the corresponding correct output, then the score would turn out to be +1. Simultaneously a loss function is declared to calculate to optimize the expected reward.

\subsection{\textbf{IRNET}~\cite{IRNETguo2019towards}}
IRNet is a neural network technique for text-to-SQL conversion that is complicated and cross-domain. IRNet solves two challenges: 1) the misalignment between the intents represented in natural language (NL) and the implementation details in SQL, and 2) the difficulty in anticipating columns due to the massive number of out-of-domain terms in the natural language. Then, using a grammar-based neural network, IRNet synthesizes a SemQL query, an intermediate form that They created to bridge the gap between NL and SQL queries. Regarding the difficult Text-to-SQL test Spider, IRNet scores 46.7 percent accuracy, which is a 19.5 percent absolute gain over the previously available state-of-the-art methods.

\subsubsection{Intermediate Representation}
The solution provided by the authors is to create a domain-specific language called SemQL that acts as an intermediary representation between NL and SQL, thereby eliminating the misalignment. Whenever a SQL query is inferred from a SemQL query, an assumption is made that the specification of a database schema is exact and comprehensive. As an illustration, consider the inference of the FROM clause in a SQL query. The algorithm begins by determining the shortest path between all the stated tables in a SemQL query in the schema.

\subsubsection{Schema Linking}
The purpose of schema linking in IRNet is to detect the columns and tables referenced in a question and assign various types to the columns based on their placement in the question.
Schema linking is an instance of entity linking in the context of Text-to-SQL, where entity refers to columns, tables, and cell values in a database.
Schema Linking starts by identifying and recognizing all of the entities stated in a question and then creating a non-overlap n-gram sequence of the question by combining those identified n-grams with the remaining 1-grams
Each n-gram in the sequence is referred to as a span, and each span is assigned a type based on the thing it represents.
\begin{figure*}[h]
    \includegraphics[width=7in]{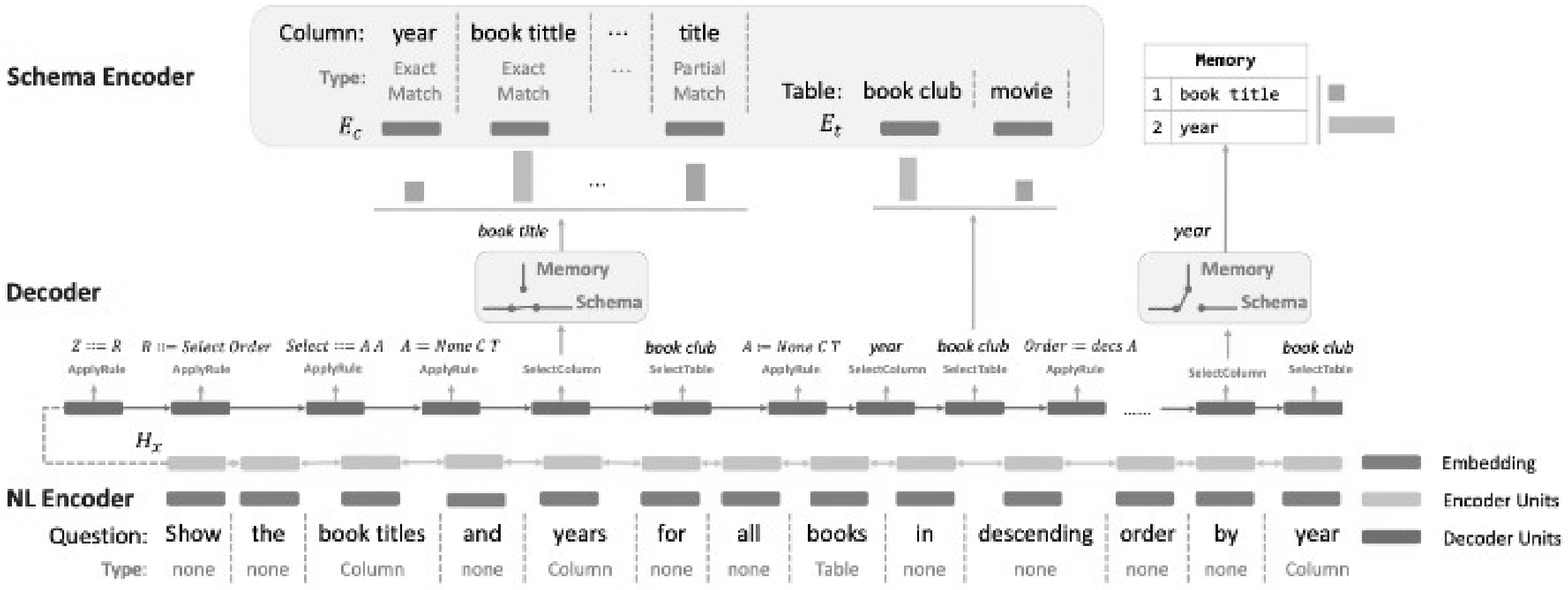}
    \caption{IRNET~\cite{IRNETguo2019towards}}
\label{arimaresults}
\end{figure*}

\subsubsection{NL Encoder}
 The NL encoder accepts an input of x and encodes it into a sequence of hidden states denoted by the letter Hx. It is transformed into an embedding vector for each word in xi, and each word's type (in xi) is also made into an embedding vector. Then, as the span embedding e I x, the NL encoder takes the average of the type and word embeddings and uses it as the span embedding. Finally, the NL encoder applies a bi-directional LSTM to all the span embeddings. When the forward and backward LSTMs are finished, the hidden output states are concatenated to form Hx.

\subsubsection{Schema Encoder}
Input data is sent to the schema encoder, producing output representing columns Ec and tables, Et. Except for the fact that They do not provide a type to a table during schema linking, the generation of table representations is identical. Each word in ci is first turned into its embedding vector, then type I is translated into an embedding vector of its own, as shown in the following diagram. Next, the schema encoder uses the mean of all word embeddings as the starting point for the column's initial representations (e). A second step is taken by the schema encoder, which is attention over the span embeddings, which results in the generation of a context vector called cic. Last, the schema encoder computes the column representation eic by adding together the initial embedding, the context vector, and the type embedding. The representations for column ci are calculated in the following manner.

\subsubsection{Decoder}
The decoder's primary objective is synthesizing SemQL queries from their constituent parts. To describe the creation process of a SemQL query through sequential applications of actions, they use a grammar-based decoder that utilizes an LSTM and is based on the tree structure of SemQL. It is possible to formalize the generating process of a SemQL query y in the following manner. Where ai is an action performed at time step i. a<i is the sequence of actions performed before I, and t is the total number of time steps spent performing the whole action sequence.

\subsection{\textbf{Model Based Interactive Semantic Parsing}~\cite{yao2019modelbasedsemanticpars}}
A model-based intelligent agent has been proposed in this paper in which an agent takes percept as the world model. In this, the parser decides when and where the interactive input is needed by adding an explanation for the same. They demonstrate two text-to-SQL datasets, which are WikiSQL and Spider. Despite lesser use intervention, this model scores a higher accuracy. 

\begin{figure}[h]
    \includegraphics[width=3.4in]{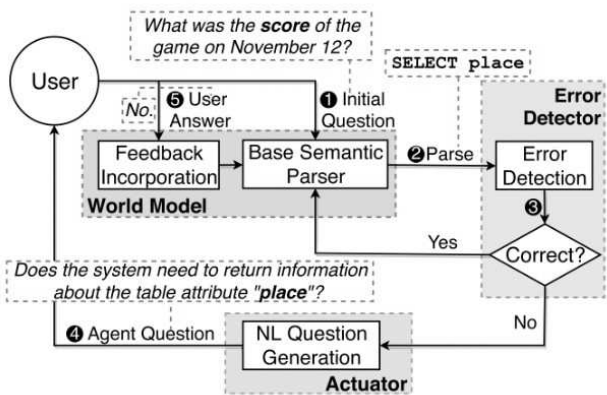}
    \caption{Model Based Interactive Semantic Parsing~\cite{yao2019modelbasedsemanticpars}}
\label{arimaresults}
\end{figure}

\subsubsection{Agent State}
Agent state st is defined as a partial SQL query, for instance, st={o1, o2, ..., ot}, where it is the predicted SQL component at time step t, such as SELECT place. 

\subsubsection{Environment}
The environment consists of a user with a purpose, which correlates to a semantic parse that the user expects the agent to provide in response to the user's intent.

\subsubsection{World Model}
Among the essential components of a MISP agent is its world model, which compacts the past perceptions throughout the interaction and forecasts the future based on the agent's knowledge of the surrounding environment.

\subsubsection{Error Detector}
The job of this module is to detect errors which it does introspective and greedy way. Two uncertainty measures are experimented with in the error detector.

\subsubsection{Actuator, a natural language generator}
An actuator records the actions of the agent through a user interface.

Given that the MISP-SQL agent executes its function by asking users binary questions, the actuator is also known as a natural language generator. A Natural language generator based on rules is defined, consisting of a seed lexicon and a grammar for generating questions from the seeds. The seed lexicon specifies an essential SQL element. Regarding MISP-SQL, there are four syntactic categories to consider: AGG for aggregates, OP for operators, COL for columns, and Q for produced queries.

The grammar describes the rules that must be followed to generate questions. Each column is detailed in its own right (i.e., the column name). The rules connected with each Q-typed item create an NL question. The Clause provides the required framework for asking meaningful questions in a meaningful way.

\subsubsection{World Model}
The agent takes into account user input and adjusts its state by a model of the world. The MISP-SQL agent leverages the basic semantic parser to transition states, saving the user time and money by eliminating the need for further training. The agent asks the user a binary inquiry regarding whether an anticipated SQL component exists. Using the NL question generator, a Q-typed item is used to generate an NL question regarding the aggregator max in the clause "SELECT max(age)".
The response either confirms or disproves the forecast made.

\subsection{\textbf{Bridging Textual and Tabular Data for Cross-Domain Text-to-SQL Semantic Parsing}~\cite{lin2020bridgingtextable}}

The authors introduce BRIDGE, an architecture bridging dependencies between natural language questions and relational databases in cross-DB semantic parsing to represent dependencies between natural language questions and relational databases. A pointer generation decoder combined with schema-consistency-driven search space pruning enabled BRIDGE to achieve state-of-the-art performance on two popular text-to-SQL benchmarks: Spider (71.1\% development and 67.5 percent test with ensemble model) and WikiSQL (71.1\% development and 67.5 percent test with ensemble model) (92.6 percent dev, 91.9 percent test). Their investigation demonstrates that BRIDGE efficiently captures the intended cross-modal interdependence and can generalize to other text-DB-related jobs.
\begin{figure*}[h]
    \includegraphics[width=7in]{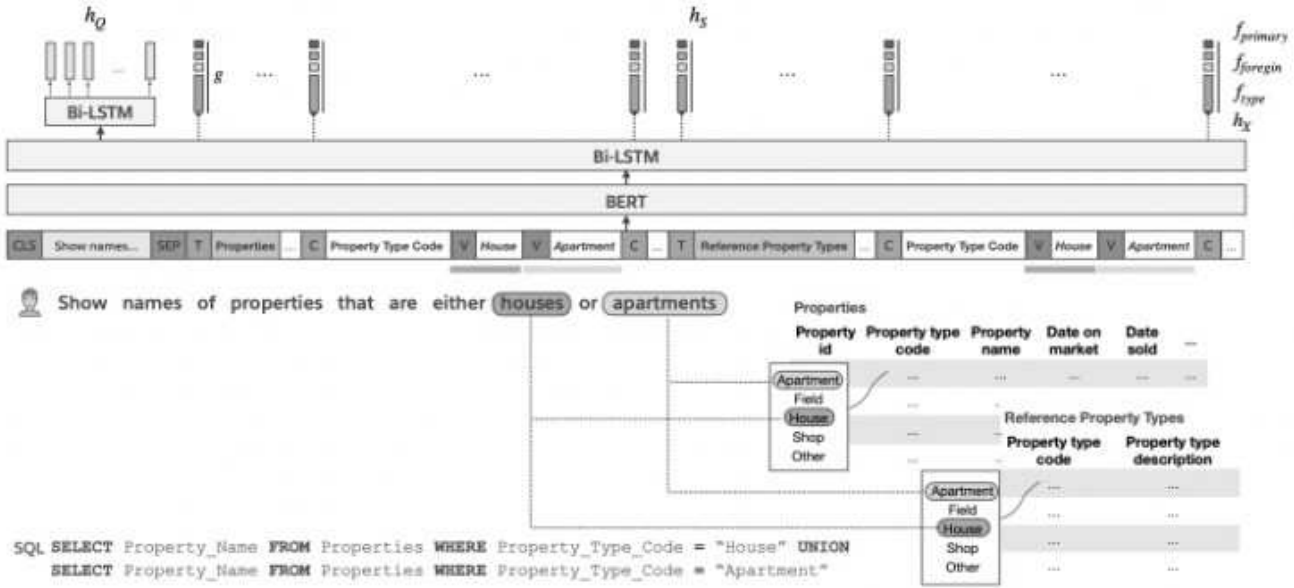}
    \caption{Bridging Textual and Tabular Data for Cross-Domain Text-to-SQL Semantic Parsing~\cite{lin2020bridgingtextable}}
\label{arimaresults}
\end{figure*}

The following is a formal definition of their cross-DB text-to-SQL job. Given a natural language question Q and the relational database schema S = hT, Ci, the parser must construct the SQL query Y that corresponds to the natural language question Q. In the schema, there are three types of tables: T1 (the first one), T2 (the second one), and T3 (the third one). There are also three types of fields: C (the last one), T1 (the first one), T2 (the second one), and T3 (the third one). A textual name is assigned to each table ti and each field cij.

\subsubsection{Question-Schema Serialisation and Encoding}
As seen in Figure~\ref{arimaresults}, each table is represented by its table name followed by the fields that make up that table. To the left of each table, the name is the unique token [T], and to the right of each field, the name is the unique token [C]. Several tables’ representations are concatenated to the question to produce a serialization of the schema. The serialization of the schema is surrounded by two [SEP] tokens and concatenated to the query. Finally, by the BERT input format, the question is preceded by [CLS] to form the hybrid question-schema serialization format.

\subsubsection{Bridging}
A fuzzy string match between Q and the picklist of each field in the DB to determine which field is correct. The matching field values (anchor texts) are entered into the question-schema representation X after the respective field names and separated by the special character [V], in the same order as the corresponding field names. If more than one value were found to match for a single field, they would concatenate all values in the matching order. The question is considered answered if a question is compared with values in many fields. They combine all matches and allow the model to learn how to resolve ambiguity.

\subsubsection{Decoder} Authors have used an LSTM-based pointer-generator in conjunction with multi-head attention. The decoder uses the question encoder’s end state as a point of departure. When the decoder reaches a new stage, it executes one of the following actions: creating a new token from the vocabulary V, copying a new token from the question Q, or copying an element from the schema S. Mathematically, at each step t, given the decoder state and the encoder representation they compute the multi-head attention as define.

\subsubsection{Schema-Consistency Guided Decoding}
Using SQL syntax limitations and the fact that the DB fields appearing in each SQL clause must only originate from the tables specified in the FROM clause, they present simple heuristics for trimming the search space of the sequence decoders.

\subsection{\textbf{Encoding Database Schemas with Relation-Aware Self-Attention for Text-to-SQL Parsers}~\cite{shin2019encodingdatabaseschemaselfatt}}
The authors’ goal is to have approaches generalizable to domains and database schemas other than the training set while converting natural language inquiries into SQL queries to answer questions from a database. For a neural encoder-decoder paradigm to effectively handle complicated questions and database schemas, it is essential to appropriately encode the schema as part of the input along with the query. Specifically, they use relation-aware self-attention inside the encoder to allow it to reason about the relationships between the tables and columns in the supplied schema and utilize this knowledge in understanding the query. On the newly released Spider dataset, they obtained considerable improvements in exact match accuracy, with 42.94 percent precise match accuracy, compared to the 18.96 percent reported in previous work.

\subsubsection{Encoding the Schema as a Graph}
The authors have started by constructing a directed graph G that represents the database schema and labels each node and edge on the graph to create an encoder. This enables reasoning about links between schema components in the encoder. The above figure illustrates the example graph. 

\subsubsection{Initial Encoding of the Input}
After that, an initial representation is extracted for each node in the graph and an initial representation for each of the words in the input query. A bidirectional LSTM over the words included in the label is used as a node label for the graph. To create the embedding for the node, they concatenate the output of the start and end time steps of this LSTM and concatenate them together. In addition, they employ a bidirectional LSTM across the words to answer the query.

\subsubsection{Relation-Aware Self-Attention}
At this point, they would want to infuse the information contained in the schema graph into these representations. They use self-attention that is also relation aware to attain this aim.

\subsubsection{Decoder}
After obtaining an encoding of the input, the method utilizes a decoder developed by Yin and Neubig [1] to construct the SQL query for the data. It constructs the SQL query as an abstract syntax tree in depth-first traversal order by generating a series of production rules that extend the last created node in the tree, as shown in the following diagram. Because the decoder is limited to selecting only syntactically valid production rules, it always delivers syntactically correct outputs. The following adjustments have been made to Yin and Neubig~\cite{yin2017syntactic} to conserve readers’ space.
\begin{itemize}
    \item When a column is to be given as output by the decoder, a pointer network~\cite{vinyals2015pointernetwork} developed on scaled dot-product attention which points to c\_final\_i and t\_final\_i  is used.
    \item The decoder accesses the encoder outputs at each step which are c\_final\_i, t\_final\_i, and q\_final\_i using multi-head attention. The original decoder in Yin and Neubig~\cite{yin2017syntactic} uses a simpler form of attention.
\end{itemize}

\subsection{\textbf{SeaD: Schema-aware Denoising}~\cite{xuan2021sead}}

Based on the Transformer design, the authors examine the questions in this study. Instead of constructing additional modules or imposing constraints on model output, they offer schema-aware denoising goals that are trained concurrently with the original S2S task. These denoising aims deal with the inherent property of logical form and, as a result, ease the schema linking necessary for the text-to-SQL operation. Erosion is used in the S2S job that trains a model to create a corrupted SQL sequence from NL and eroded schema. These suggested denoising goals and the origin S2S job are combined to train a SeaD model. In addition, to overcome the limitations of execution-guided (EG) decoding, the authors present a clause-sensitive EG technique that determines beam size based on the projected clause token. The findings demonstrate that Their model outperforms earlier work and provides a new benchmark for WikiSQL. It reflects the efficiency of the schema-aware denoising technique and highlights the significance of the task-oriented denoising aim.

They generate the SQL sequence token by token using auto-aggressive generation. The transformer is a widely used S2S translation and generation architecture. In this part, they first give an example formulation that transforms text-to-SQL into a standard S2S work, then They introduce the Transformer architecture with a pointer generator. Then They discuss schema-aware denoising and clause-sensitive EG decoding.

\subsubsection{Sample Formulation}
S2S creation requires re-formatting the structural target sequence and unordered schema set. Each schema column name is preceded with a specific token, where I signify the i-th column. The column type is also included in the name sequence [col name]: [col type]. The representative sequence for the schema is obtained by concatenating all columns in the schema. For model input, the schema sequence is combined with the NL sequence. They initiate SQL sequence with raw SQL query and modify it: 1) Surrounding SQL entities and values with a "`" token and deleting other tokens; 2) Replacing col entities with their matching schema token; 3) Inserting spaces between punctuation and words.

\subsubsection{Transformer with Pointer}

In the decoding procedure, each input sequence Isource is encoded using the transformer encoder into the hidden states Htarget. First, the transformer decoder generates the hidden states ht in step t based on the previously generated sequence and encoded output. Next, an affine transformation is applied to obtain scores.

\subsubsection{Schema Aware Denoising}

Two schema-aware objectives are proposed, namely, erosion and shuffle, that train the model to either rebuild the original sequence from applying noise to the input or otherwise predict the corrupted output.



\subsubsection{Clause-sensitive EG Decoding}
The projected SQL may include mistakes resulting from improper schema linking or grammar during text-to-SQL inference. Through an executor-in-loop iteration, EG decoding is suggested to rectify these problems. It is carried out by sequentially feeding SQL queries from the candidate list to the executor and eliminating those that fail to execute or produce an empty response. Such a decoding technique, although successful, shows that the primary dispute in the candidate list is centered on schema linking or grammar. They are directly applying EG to the candidates obtained by beam search results in a negligible improvement since they consist of redundant variants with selection or schema naming as the primary emphasis, etc. This issue may be resolved by changing the beam length of most expected tokens to 1 and releasing tokens associated with schema linking (e.g., WHERE).

\subsection{\textbf{Learning to Synthesize Data for Semantic Parsing}~\cite{wang2021learningsynthesizedata}}

The authors utilize text-to-SQL as an example problem and propose a generative model to generate pair-wise utterance-SQL representations. Learning to Synthesize Data for Semantic Parsing, the authors first simulate the distribution of SQL queries using a probabilistic context-free grammar (PCFG). Then, with the aid of a SQL-to-text translation model, the relevant SQL query expressions are constructed. In their scenario, the 'target language' is a formal language whose underlying grammar is well understood. Like the training of a semantic parser, data synthesizer training needs a collection of utterance-SQL pairings. Their two-stage data synthesis strategy, consisting of the PCFG and the translation model, is more sample efficient than a neural semantic parser, which is very similar to back-translation~\cite{sennrich-etal-2016-improving}. They sample synthetic data from the generative model to train a semantic parser.

\subsection{\textbf{Awakening Latent Grounding from Pretrained Language Models for Semantic Parsing}~\cite{liu2021awakening}}
The authors propose the new Erasing-then-Awakening method (ETA). Recent developments influence it in interpretable machine learning~\cite{kumar2020challenges}, where the significance of individual pixels may be evaluated in the classification decision. Similarly, their method first assesses the importance of each word to each idea by deleting it and examining the variance of concept prediction judgments (elaborated later). Then, it leverages these contributions as pseudo labels to reawaken PLMs' dormant grounding. In contrast to previous research, their technique requires supervision of concept prediction, which may be readily extracted by downstream activities (e.g., text-to-SQL). Four empirical datasets illustrate that their method may uncover latent grounding that is understandable to human experts. In training, their method is not subjected to any human-annotated grounding label. Hence this problem is very complex. Significantly, they discover that the grounding may be readily connected with downstream models to increase their performance by as much as 9.8 percent.

The author’s model comprises PLM, CP, and grounding modules. This section begins with an overview of ETA's training program, which consists of three steps: (1) Train an auxiliary module for idea prediction. (2) Remove tokens from a question to derive concept prediction confidence differences as pseudo alignment. (3) Awaken dormant grounding in PLMs by supervising them with pseudo alignment. Then, the process for producing grounding pairs in inference is introduced.

\subsection{\textbf{SyntaxSQLNet}~\cite{yu2018syntaxsqlnet}}
A syntax tree network to deal with the complex and cross-domain task of text-to-SQL generation SyntaxSQLNet can handle a substantially higher number of sophisticated SQL examples (such as which can handle nested queries on unknown databases) than earlier research, beating the previous state-of-the-art model in precise matching accuracy by 7.3 \%. SyntaxSQLNet may enhance speed by an additional 7.5 \%utilizing a cross-domain augmentation strategy, for a total improvement of 14.8 \%. On specific complicated text-to-SQL benchmarks, such as ATIS and GeoQuery, Seq2Seq encoder-decoder designs may achieve more than 80\% accurate matching accuracy. Although these models appear to have addressed the majority of challenges in this field, most learn to match semantic parsing outcomes rather than genuinely understanding the meanings of inputs.

SyntaxSQLNet is a SQL-specific syntax tree network that may be used to solve the Spider problem. Researchers build a syntax tree-based decoder with SQL generation path history to construct complicated SQL queries with numerous clauses, selects, and sub-queries. They also create a table-aware column encoder to help Their model learn to generalize to new databases with new tables and columns. 

\subsubsection{Approach}
SyntaxSQLNet divides the SQL decoding process into nine modules, each responsible for predicting distinct SQL components such as keywords, operators, and variables. The IUEN Module, which predicts INTERSECT, UNION, EXCEPT, and NONE, decides whether They need to call themselves again to construct nested queries. WHERE, GROUP BY, and ORDER BY keywords are predicted by the KW Module, whereas SELECT is used in every query. COL Module is used for predicting table columns and OP Module. MAX, MIN, SUM, COUNT, AVG, and NONE are used in the AGG Module to anticipate aggregators. Predicting the ROOT of a new subquery or terminal value is done by the Root/Terminal Module. AND/OR Module predicts whether an AND or OR operator exists between two conditions. The ORDER BY keywords is predicted by the DESC/ASC/LIMIT module. It is only used when ORDER BY is anticipated first. The HAVING Module does predict the presence of HAVING for the GROUP BY clause. It is only used when GROUP BY has been expected before.

\subsubsection{SQL Grammar}
SyntaxSQLNet chooses which module to run and forecast the next SQL token to build depending on the current SQL token and the SQL history (the tokens they've gone over to arrive at the current token) during the decoding process. During decoding, the model checks the current token instance type and whether the previously decoded SQL token is GROUP for HAVING and WHERE or HAVING for OP.

\subsubsection{Input Encoder}
Each module's inputs consist of three data types: a query, a database structure, and the current SQL decoding history path. A bi-directional LSTM, BiLSTMQ, is used to encode a question sentence.

\subsubsection{Table-Aware Column Representation}
In particular, given a database, as an initial input to each column, SyntaxSQLNet obtains a list of words in the table name, words in the column name, and the type of information of the column (string, integer, primary/foreign key). The table-aware column representation of a particular column is then computed as the final hidden state of a BiLSTM executing on top of this sequence, just as SQLNet. In this manner, the encoding scheme can interpret a natural language inquiry in the context of the supplied database by capturing both global (table names) and local (column names and types) information in the database design.

Each SQLNet module, by contrast, ignores the preceding decoded SQL history. As a result, if SyntaxSQLNet had applied it straight to their recursive SQL decoding stages, each module would anticipate the same outcome each time it was called. Each module can forecast a different output based on the history each time it is called during the recursive SQL generation process by supplying the SQL history. Additionally, the SQL history can assist each module in performing better on long and complicated queries by allowing the model to record the relationships between clauses. During test decoding, predicted SQL history is utilized. To build gold SQL route history for each training sample, SyntaxSQLNet first explores each node in the gold query tree in pre-order.

\subsubsection{Recursive SQL Generation}
The SQL creation procedure involves recursively activating many modules. To structure Their decoding procedure, as shown in Figure 2, They SyntaxSQLNet uses a stack. Each decoding step consists in removing one SQL token instance from the stack, using a grammar-based module to predict the next token instance, and then pushing the anticipated instance into the stack. Decoding continues until the stack is depleted. In the first decoding phase, this method deliberately creates a stack with only ROOT. The stack then pops ROOT in the following stage. The ROOT uses the IUEN module to determine whether there is an EXCEPT, INTERSECT, or UNION. If that's the case, the next step is to create two subqueries. If the model predicts NONE, it will be placed at the bottom of the stack. At the next step, the stack pops NONE. In Figure 2, the currently popped token is SELECT, a keyword (KW) type instance. It uses the COL module to forecast the column’s name that will be pushed to the stack.

\subsubsection{Comparison to Existing Models}
Even though the individual modules are similar to SQLNet and TypeSQL, the syntax-aware decoder allows the modules to generate complex SQL queries recursively based on the SQL grammar.  This result suggests that the syntax and history information benefit this complex text-to-SQL task. Specifically, SyntaxSQLNet outperforms the previous best, SQLNet, even without the data augmentation technique, by 7.3\%.

\subsection{PointerSQL~\cite{wang2018robustpointersql}}
The PointerSQL solves the challenge of neural semantic parsing, which converts natural language inquiries into SQL queries that can be executed. The authors propose a novel approach for using SQL semantics called execution guidance. By conditioning the execution of a partially created program, it discovers and rejects erroneous programs during the decoding step. The method may be employed with any autoregressive generating model, as demonstrated by the authors using four cutting-edge recurrent or template-based semantic parsing models. They also show that execution guidance increases model performance across the board on various text-to-SQL datasets with varying sizes and query complexity, including WikiSQL, ATIS, and GeoQuery. Consequently, they reached a new state-of-the-art execution accuracy on WikiSQL of 83.8\%.

Developing effective semantic parsers to translate natural language questions into logical programs has been a long-standing goal. The PointerSQL focuses on the semantic parsing task of translating natural language queries into executable SQL programs. It shows how to condition such models to avoid whole classes of errors to generate syntactically valid queries. Execution guidance is the idea that a partially generated query can already be executed in languages such as SQL. The results of that execution can be used to guide the generation procedure. In other words, execution guidance extends standard autoregressive decoders to condition on non-differentiable partial execution results at appropriate timesteps additionally. It also shows the effectiveness of execution guidance by extending a range of existing models with execution guidance and evaluating the resulting models on various text-to-SQL tasks.

\subsubsection{Execution-Guided Decoding:} The authors of PointerSQL combine the query generation method with a SQL execution component to prevent creating queries that result in Execution Errors, and extending a model with execution guidance, therefore, necessitates deciding which stages of the generating method to execute the partial result at, and then using the outcome to modify the remaining generation procedure. The pseudocode of an execution-guided expansion of a typical autoregressive recurrent decoder is demonstrated. It's a model-specific decoder cell called DECODE, an extension of ordinary beam search. The technique keeps just the top k states in the beam that correspond to the partial programs without execution faults or empty outputs whenever possible (where the result at the current timestep t corresponds to a partial executable program).
Execution guidance can be employed as a filtering step after decoding in non-autoregressive models based on feedforward networks, for example, by removing result programs that produce execution errors. Any autoregressive decoder may be used similarly at the end of beam decoding. In many application areas (including SQL creation), execution checks may be applied to partially decoded programs rather than only after beam decoding.
For example, right after the token 'Haugar' is emitted, this helps to remove an improperly produced string-to-string inequality comparison "... WHERE opponent > 'Haugar'... " from the beam. This considerably enhances the efficacy of execution guidance, as demonstrated by the research.

Wang et al.~\cite{wangpointing} introduced the Pointer-SQL model, which extends and specializes in the sequence-to-sequence architecture of the WikiSQL dataset. As inputs, it takes a natural language question and a single table of its schema. To learn a joint representation, a bidirectional RNN with LSTM cells processes the concatenation of the table header (column names) of the queried table and the question as input. Another RNN, the decoder, can look over and copy the encoded input sequence. The decoder uses three separate output modules to correspond to three decoding types, a vital feature of this model. One module generates SQL keywords, while the other is for copying a column. Extending the Coarse2Fine model with an execution-guided decoder improves its accuracy by 5.4\% on the WikiSQL test, which beats state-of-the-art on this task.

\subsection{Content Enhanced BERT-based Text-to-SQL Generation~\cite{content_enhanced_BERT}}
BERT is a transformer-based model with an intense level of complexity. It uses the mask language model loss and the next-sentence loss to pre-train on a vast corpus. Then They could fine-tune BERT for various tasks, including text categorization, matching, and natural language inference, to achieve new state-of-the-art performance.

The NL2SQL model depends on the following three things:
\begin{enumerate}
  \item To mark the query, they take the match information from all of the table columns and the query string to create a feature vector that has the same length as the query.
  \item To mark the column, they utilize the match info of all the table column names and question strings to create a feature vector that has the same length as the table header. 
  \item The whole BERT-based model is designed with the two feature vectors above as external inputs. 
\end{enumerate}

\subsection{GRAPPA~\cite{yu2021grappa}}
The authors provide a unique grammar-augmented pre-training approach for table semantic parsing in this research (GRAPPA). The authors infer a synchronous context-free grammar (SCFG) for natural mapping language to SQL queries from existing text-to-SQL datasets that account for most question-SQL patterns. The authors may generate a question-SQL template from a text-to-SQL example by excluding references to schema components (tables and fields), values, and SQL actions. Using a unique text-schema linking goal that predicts the syntactic function of a table column in the SQL for each pair, The authors train GRAPPA on these synthetic question-SQL pairings and their related tables. To pre-train GRAPPA, the authors use 475k synthetic cases and 391.5k examples from existing table-and-language datasets. Their method significantly decreases training time and GPU costs. The authors examine four major semantic parsing benchmarks under both strong and light supervision. GRAPPA routinely obtains new state-of-the-art results on all of them, exceeding all previously published results by a wide margin.

\subsubsection{Methodology}
Increasing research demonstrates that using enhanced data does not always result in a significant performance improvement in cross-domain semantic parsing end jobs. The most obvious explanation is that models tend to overfit the canonical input distribution, producing notably distinct utterances from the originals. In addition, instead of immediately training semantic parsers on the enriched data, the author’s study is the first to employ synthetic instances in pre-training to inject an inductive compositional bias into LMs and demonstrate that it truly works if the overfitting issue is treated with care. To combat overfitting, the author’s pre-training data also contains a tiny number of table-related utterances. As a regularization element, the authors apply an MLM loss to them, which challenges the model to balance actual and synthetic cases during pre-training. This reliably enhances the performance of all semantic parsing jobs that follow.

\subsection{TRIAGESQL~\cite{zhang2020didtriagesql}}
TRIAGESQL is a benchmark for cross-domain text-to-SQL question intention categorization. The authors describe four categories of unanswerable questions and distinguish them from answerable ones, then generate a benchmark dataset consisting of 34,000 databases and 390,000 questions from 21 text-to-SQL,
question answering, and table-to-text existing datasets. The authors reviewed and annotated data to create a high-quality test set with 500 instances of each kind. The fact that a Transformer-based model RoBERTa~\cite{liu2019roberta} got a 60 percent F1 score on their benchmark for a five-class classification model demonstrates the difficulty of this endeavor.

\subsection{ValueNet~\cite{brunner2021valuenet}}
Brunner et al~\cite{brunner2021valuenet} proposed two end-to-end texts to SQL algorithm ValueNet and ValueNet light. The idea of Valuenet is based on combining the metadata information from the database with the information on base data from the spider dataset. The ValueNet is based on encode-decoder architecture to generate the SQL query by extracting values from a user question which helps them to generate other possible value candidates which are not used or mentioned in the question asked in the form of natural language text.

ValueNet light selects the correct values from a given list of possible ground truth values and then synthesizes a full query including the chosen values. Whereas ValueNet goes one step ahead with extracting and generating value candidates from the basic question format in the natural language and the content from the database only using Name Entity Recognition (NER) and heuristics. The next step uses those value candidates to generate SQL queries using encoder-decoder. However, between ValueNet and ValueNet light, a performance gap of $3\%-4\%$ is expected.

\subsection{RYANSQL~\cite{choi2021ryansql}}

RYANSQL (Recursively Yielding Annotation Network for SQL) is another approach to solving cross-domain complex Text-to-SQL problems by generating multiple nested queries and recursively predicting its component SELECT statements using Statement Position Code (SPC) and sketch-based slot filling. RYANSQL improves the previous state-of-the-art system by 3.2\% in terms of the exact matching accuracy test using BERT.
The two simple input manipulations are used to improve the performance of RYANSQL. One is that some tables are used only to make a “link” between other tables in a FROM clause. And in the second manipulation, the table names are supplemented with their column names which help the architecture to distinguish between the same column names but from different tables.

\subsection{F-SQL~\cite{zhang2020f-sql}}

F-SQL focus on solving the problem of table content utilization. It employs the gate mechanism to fuse table schemas and contents and get a different representation of table schemas. It uses sketch-based~\cite{xu2018sqlnet} technique to synthesize SQL queries from natural language questions.

Since sketch-based techniques require neural network models to predict the slots to assemble SQL, F-SQL uses multiple sub-models in its architecture to predict the slot in the sketch. The selection of sub-models also depends on the dataset and the pre-trained encoder. They achieved better SQL query synthesis by training all the slots together. To overcome the challenge of column prediction, they used a simple gate mechanism to fuse table schemas and table contents and get were able to predict distinct columns.

%% file: sections/conclusion.tex
\section{Conclusion}
In this profound work-in-progress study, besides the overview of various approaches and methods for TEXT to SQL language models, we also focus on terminologies and definitions frequently used in their summarized form for the reader to gain deep insight into natural language processing. This was further stratified based on their domain of application. Proceeding, upon extensive research and study, a variety of union of benchmarks were discovered. Adding to that, a multitude of models were described in detail. The stance toward individual models has been categorized based on a broad division of approaches. This allows a simplified overview of the evolution of a specific approach. Following this, a detailed summary of the multiple experiments, evaluation, and accuracy of the models mentioned above. In our future study, we want to cover all the mentioned TEXT2SQL approaches in full detail with our own set of benchmark datasets. This will also help to have a standard benchmark of all the models on a similar benchmark dataset which is not the case currently. This will also help in accessing the applicability of specific models and their limitations on different datasets.